\documentclass[letterpaper]{article} 
\usepackage{aaai2026}  
\usepackage{times}  
\usepackage{helvet}  
\usepackage{courier}  
\usepackage[hyphens]{url}  
\usepackage{graphicx} 
\usepackage{natbib}  
\usepackage{caption} 
\usepackage{algorithm}
\usepackage{algorithmic}
\usepackage{amssymb}
\usepackage{booktabs} 
\usepackage{multirow} 
\usepackage{amssymb}  
\usepackage{amsmath}
\usepackage{graphicx}
\usepackage{subcaption}
\usepackage{pifont} 
\def\ie{$i.e.$}
\usepackage{newfloat}
\usepackage{listings}
\DeclareCaptionStyle{ruled}{labelfont=normalfont,labelsep=colon,strut=off} 
\floatstyle{ruled}
\newfloat{listing}{tb}{lst}{}
\floatname{listing}{Listing}
\title{HSA-Net: Hierarchical and Structure-Aware Framework for Efficient and Scalable Molecular Language Modeling}
\author{
Zihang Shao\textsuperscript{\rm 1}, 
Wentao Lei\textsuperscript{\rm 1}%
\thanks{Equal Contribution.}
Lei Wang\textsuperscript{\rm 2}, 
Wencai Ye\textsuperscript{\rm 2}, 
Li Liu\textsuperscript{\rm 1}%
\thanks{Corresponds to Li Liu (avrillliu@hkust-gz.edu.cn)}
}

\affiliations{
\textsuperscript{\rm 1}The Hong Kong University of Science and Technology (Guangzhou) \\
\textsuperscript{\rm 2}Jinan University \\
}

\usepackage{bibentry}
\nocopyright
\begin{document}

\maketitle

\begin{abstract}
Molecular representation learning, a cornerstone for downstream tasks like molecular captioning and molecular property prediction, heavily relies on Graph Neural Networks (GNN). However, GNN suffers from the over-smoothing problem, where node-level features collapse in deep GNN layers. While existing feature projection methods with cross-attention have been introduced to mitigate this issue, they still perform poorly in deep features. This motivated our exploration of using Mamba as an alternative projector for its ability to handle complex sequences. However, we observe that while Mamba excels at preserving global topological information from deep layers, it neglects fine-grained details in shallow layers. The capabilities of Mamba and cross-attention exhibit a global-local trade-off. To resolve this critical global-local trade-off, we propose \textbf{H}ierarchical and \textbf{S}tructure-\textbf{A}ware Network (\textbf{HSA-Net}), a novel framework with two modules that enables a hierarchical feature projection and fusion. Firstly, a \textbf{H}ierarchical \textbf{A}daptive \textbf{P}rojector (\textbf{HAP}) module is introduced to process features from different graph layers. It learns to dynamically switch between a cross-attention projector for shallow layers and a structure-aware Graph-Mamba projector for deep layers, producing high-quality, multi-level features. Secondly, to adaptively merge these multi-level features, we design a \textbf{S}ource-\textbf{A}ware \textbf{F}usion (\textbf{SAF}) module, which flexibly selects fusion experts based on the characteristics of the aggregation features, ensuring a precise and effective final representation fusion. Extensive experiments demonstrate that our HSA-Net framework quantitatively and qualitatively outperforms current state-of-the-art (SOTA) methods. 
\end{abstract}
\noindent\textbf{Demo page} — \url{https://hsa-net.github.io/}
\section{Introduction}
Molecular representation learning is a cornerstone of modern computational chemistry, crucial for generating expressive features that satisfy downstream tasks like molecular captioning \cite{edwards2022translation} and molecular property prediction \cite{zhao2025dual}. 

\begin{figure}[t]
    \centering
    \includegraphics[width=1.0\columnwidth]{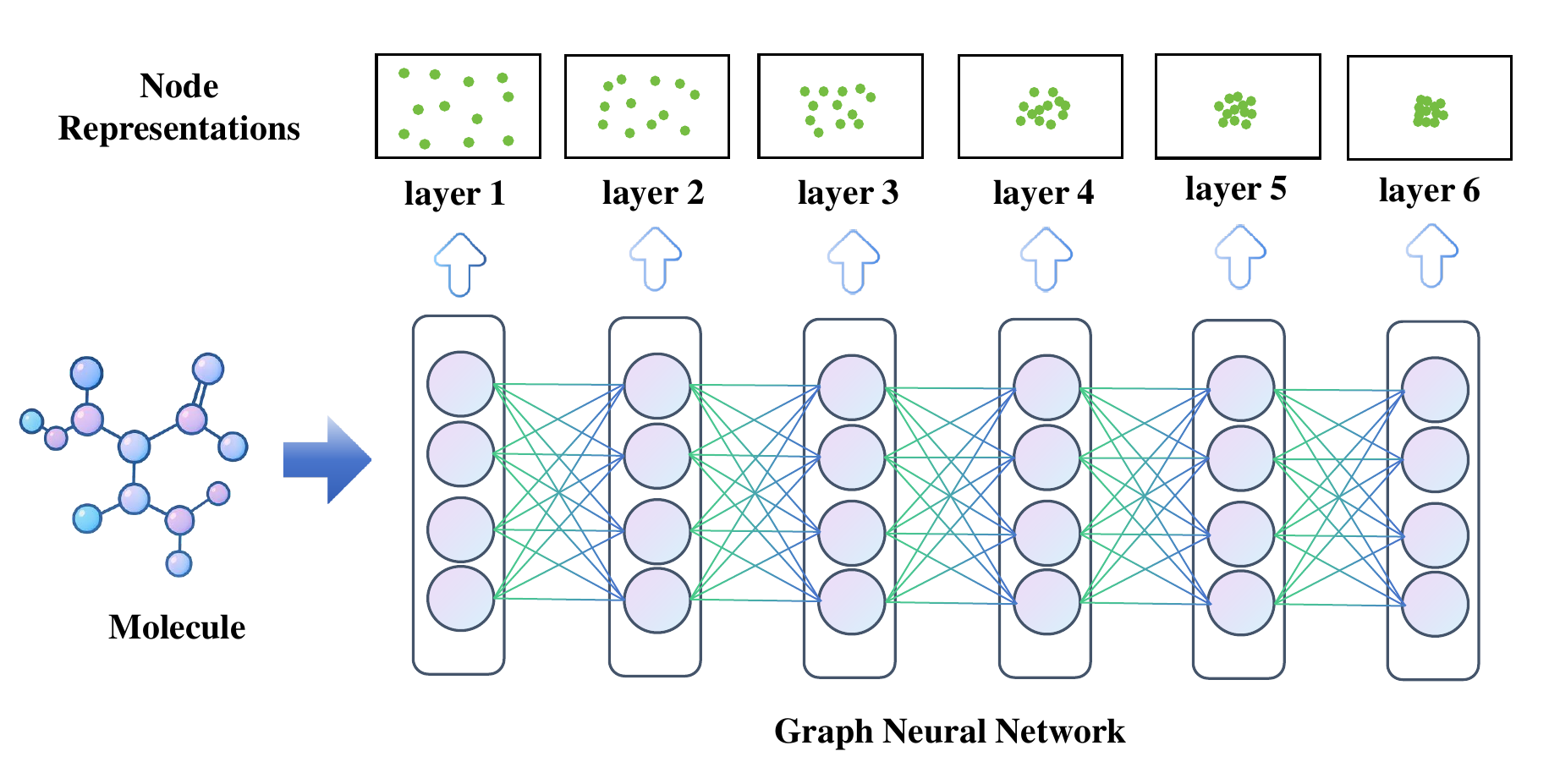} 
    \caption{The illustration of GNN over-smoothing problem in molecular representation learning. As the layers become deeper, node representations collapse into uniform features. }
    \label{fig:over}
\end{figure}

Previous work on molecular representation learning is mainly based on Graph Neural Networks (GNN) \cite{scarselli2008graph}, which iteratively aggregate feature information from local neighborhoods to learn expressive node representations. To capture global patterns within a graph, a common strategy is to stack multiple GNN layers. However, this approach is fundamentally constrained by the over-smoothing problem \cite{li2018deeper}. As shown in Figure \ref{fig:over}, information propagates through GNN layers and the iterative message-passing causes distinct node-level representations to collapse into uniform, non-informative features, which neglects the structural details that differentiate molecules. Previous work \cite{park2024llamo} has empirically proven that introducing a multi-layer cross-attention feature projection method can alleviate the over-smoothing problem. Although the feature projection approach can reduce over-smoothing, it may not completely eliminate it. The representations from deep layers might still be highly smoothed, and the existing projector cannot extract this information very well.




\begin{figure}[h]
    \centering
    \includegraphics[width=1.0\columnwidth]{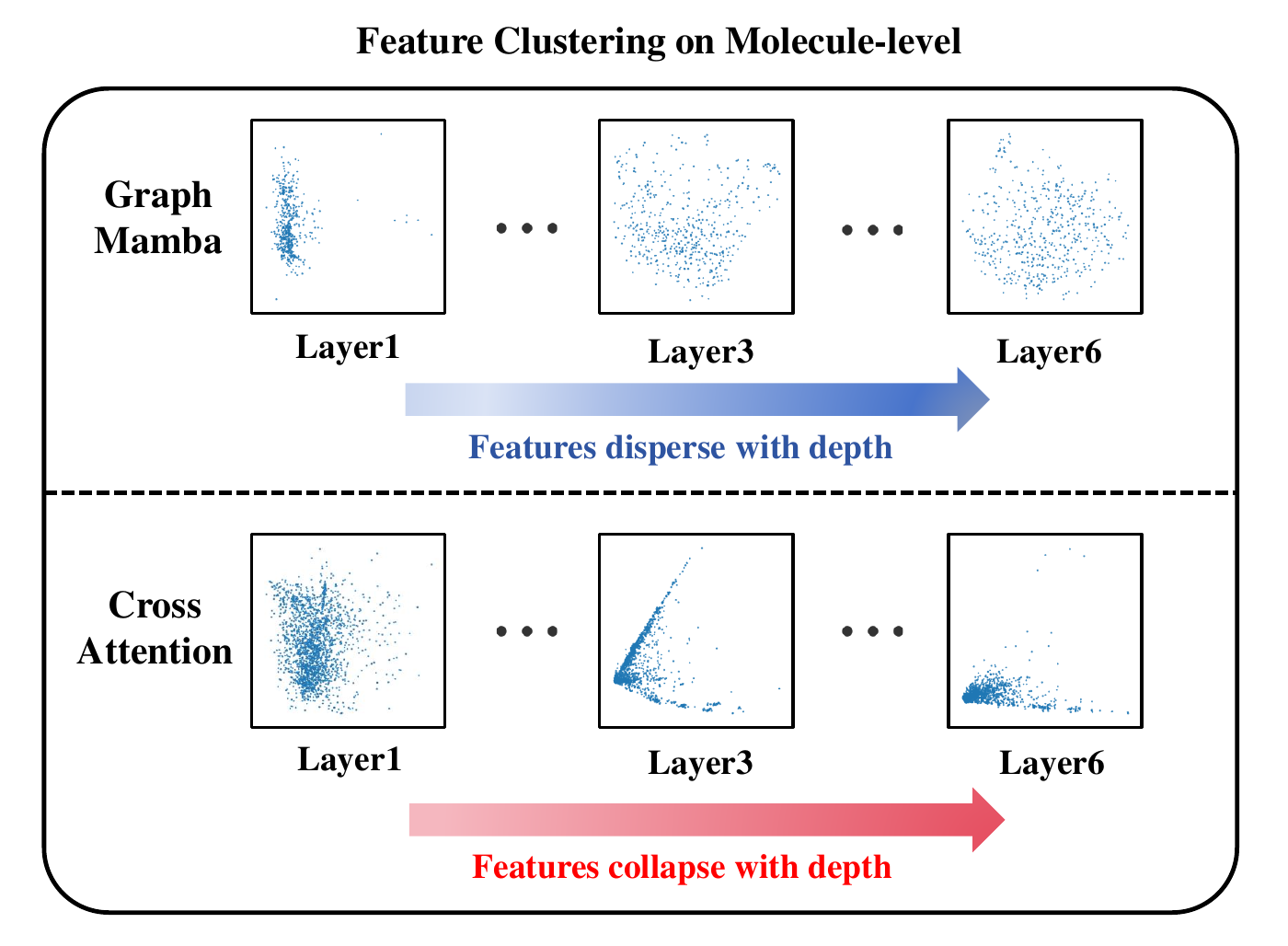} 
    \caption{Visualization of molecule feature distributions using t-SNE. The plots compare the final graph-level representations of a molecule dataset at GNN Layer 1 and Layer 6, generated by Cross-Attention and Graph-Mamba.}
    \label{fig:collapse}
\end{figure}

Recently, mamba \cite{gu2023mamba} have emerged as an effective alternative for sequential feature processing. Introducing mamba as a feature projector is a choice worth exploring. After conducting the primary experiment, we have an enlightening discovery: We find that mamba and cross attention, as two different projectors, show two opposite and complementary trends for different layers in the molecular feature projection task. Note that our experiment was conducted at the molecule-level because we believed that the node-level feature is insufficient to reflect the impact of feature degradation on down-stream tasks. Specifically, we conducted a visualization of molecule-level feature representations generated by these two distinct methods. Our analysis, depicted in Figure \ref{fig:collapse}, reveals that both methods exhibit a form of representation collapse, but in opposing manners. For Cross-Attention, although feature projection is conducted, the features still tend to collapse from a dispersed state in shallow layers to a highly concentrated cluster in deep layers. Conversely, Graph-Mamba displays an opposite trend: its features evolve from a relatively clustered state in shallow layers to a more dispersed distribution in deep layers. Based on above observation, cross-attention is more suitable for handling shallow features, which contains more fine-grained local information (\ie, chemical bonds), while mamba is more suitable for handling deep features, which can reflect structural and global information. Overall, both methods perform well in some layers but exhibit a global-local trade-off.

Based on the above insight, we propose the \textbf{H}ierarchical and \textbf{S}tructure-\textbf{A}ware Network (HSA-Net) which combines the capabilities of these two projectors to resolve this critical global-local trade-off. HSA-Net learns expressive molecular representations by dynamically adjusting its feature projection methods to mitigate the hierarchical degradation of information. 

Specifically, HSA-Net has two modules that enable hierarchical feature projection and fusion: Firstly, the Hierarchical Adaptive Projector (HAP) module is designed to process features from different GNN layers. The HAP learns to dynamically switch between two feature projectors: a cross-attention projector, which is effective for identifying fine-grained patterns in shallow layers, and a structure-aware Graph-Mamba projector, which excels at preserving feature structure information and modeling sequential context in deep layers. This dynamic allocation effectively produces high-quality, diverse features while mitigating the global-local trade-off in feature projection.
Secondly, to adaptively assemble the diverse features produced by the HAP, we design the Source-Aware Fusion (SAF) Module. This module at the final stage of the network flexibly selects different fusion experts based on the characteristics of the features, ensuring a precise and effective final representation fusion.

Our contributions can be summarized as follows:
\begin{itemize}
\item{We propose HSA-Net, a novel framework that introduces a hierarchical adaptive feature projection and fusion method to molecular representation learning, resolving the global and local trade-off in molecule feature projection.}
\item{Two novel modules are designed: 1) HAP module is designed to adaptively select projector to achieve feature projection from different GNN layers. 2) SAF module, which use a MOE mechanism to adaptively merge multi-source information.}
\item{Extensive experiments are conducted on six public datasets for three down-stream tasks: Molecule Description, IUPAC Prediction and Property Prediction. Our proposed HSA-Net quantitatively and qualitatively outperforms current SOTA methods. Additional ablation experiments and visualization results are also provided to prove the effectiveness of our method.}
\end{itemize}

\section{Related Work}

\subsection{Molecular Representation Learning}
Molecular representation learning aims to encode complex chemical structures into low-dimensional vectors that are suitable for downstream tasks. Early approaches were feature-driven, relying on handcrafted molecular fingerprints like ECFP \cite{rogers2010extended} and descriptors that quantify chemistry properties. These methods often fail to capture the context of a molecule's graph structure.

The advance of GNNs marked a paradigm shift, enabling models to learn representations directly from the molecular graph. Architectures like GCN \cite{kipf2016semi}, GAT \cite{velivckovic2017graph}, and DMPNN \cite{yang2021deep} utilize message-passing to iteratively aggregate information from neighboring nodes, proving highly effective for various property prediction tasks. However, a key limitation of standard GNNs is the over-smoothing phenomenon. As network depth increases to expand the receptive field, node representations tend to become indistinguishable, leading to a loss of critical local structural information. This problem fundamentally limits the ability of deep GNNs to model global topological patterns without sacrificing local detail.

\begin{figure*}[h]
    \centering
    \includegraphics[width=2.1\columnwidth]{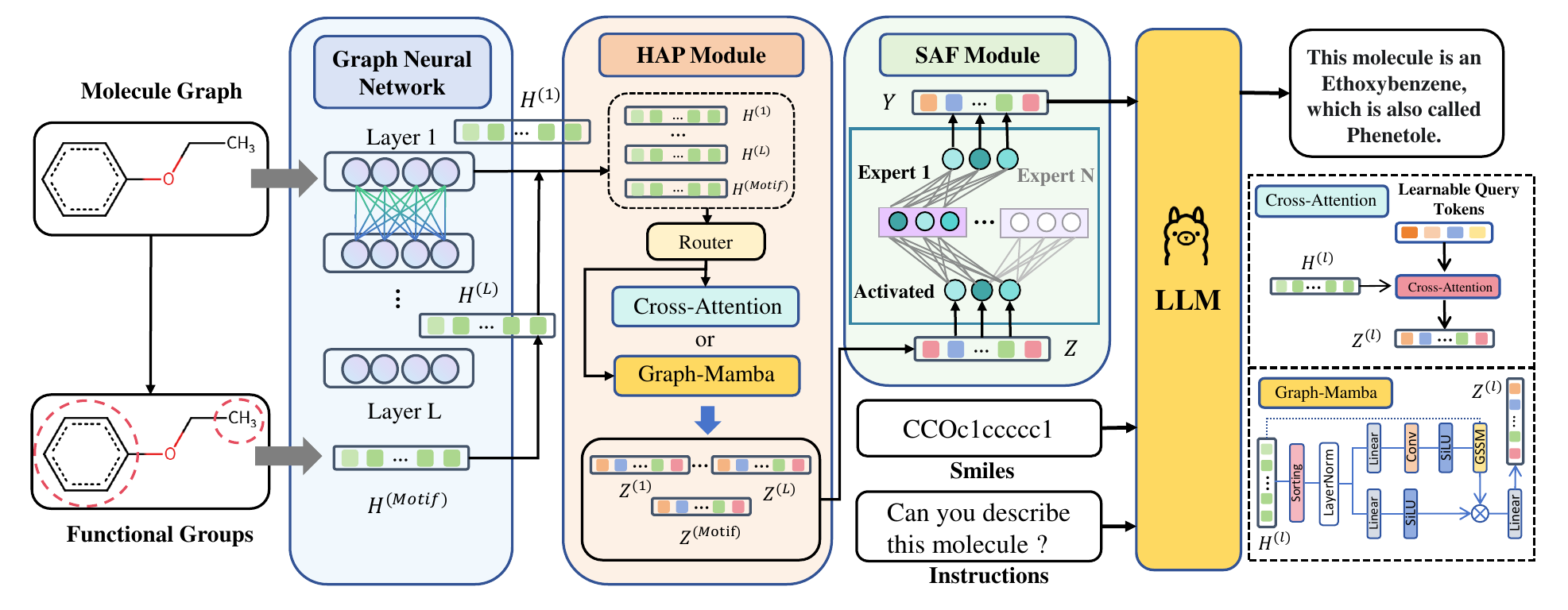} 
    \caption{The overall architecture of the HSA-Net framework. It consists of a GNN encoder, a HAP module to process multi-level graph features, and a SAF module to integrate multi-source information into a final, expressive representation.}
    \label{fig:architecture}
\end{figure*}

\subsection{Mambas for Molecule Learning}

Recently, state-space models (SSMs) like Mamba \cite{gu2023mamba} have emerged as an effective alternative for sequence modeling, demonstrating linear-time complexity and strong performance on long-range dependency tasks. Their application to graphs \cite{hu2025mol}, has shown promise in capturing sequential context within graph structures. Our work leverages this capability, positioning Graph-Mamba as a dedicated structure-aware projector to complement the global perspective of cross-attention.


\subsection{Integrating Molecular Graphs with LLM}
Recent advances in large language models (LLMs), such as Galactica \cite{taylor2022galactica}, GPT-4 \cite{achiam2023gpt}, and LLaMA \cite{touvron2023llama}, have enabled notable progress in molecular-language tasks including captioning, IUPAC naming, and property prediction.    To leverage the strengths of LLMs, several methods integrate graph-based molecular features via MLP projections \cite{park2024llamo} or cross-modal projectors \cite{tran2025xmolcap}.  However, these projection strategies still fail to simultaneously capture local structural details and global topological awareness, while also inheriting limitations of GNN encoders, such as the over-smoothing problem.

\section{Methodology}

To resolve the trade-off between capturing global patterns and preserving local structural details, we introduce a novel architecture with a two-stage framework to effectively to align molecular features across multiple levels with downstream tasks. This framework comprises two key modules: HAP, which adaptively selects optimal feature projector for hierarchical inputs, and SAF module, which adaptively integrates the final heterogeneous representations. The overall architecture is shown in Figure \ref{fig:architecture}.

\subsection{Graph Encoder and Hierarchical Features}
Given a molecule, we first represent it as a graph $\mathcal{G} = (\mathcal{V}, \mathcal{E})$, where $\mathcal{V}$ is the set of atoms (nodes) and $\mathcal{E}$ is the set of bonds (edges). We employ a standard $L$-layer GNN as a graph encoder to extract node features at different hierarchical levels. The output of the $l$-th layer, denoted as $\mathbf{H}^{(l)} \in \mathbb{R}^{|\mathcal{V}| \times d}$, captures the structural information within an $l$-hop neighborhood of each node.
\begin{equation}
    \mathbf{H}^{(l)} = \text{GNN}^{(l)}(\mathbf{H}^{(l-1)}, \mathcal{E}) .
\end{equation}
The collection of these multi-level features, $\{\mathbf{H}^{(1)}, \mathbf{H}^{(2)}, \dots, \mathbf{H}^{(L)}\}$, represents the hierarchical information of the molecule, which serves as the primary input to our HAP module.

To further capture high-level molecular semantics, we introduce a dedicated module for processing molecular motifs, inspired by \cite{ji2022relmole}. In this work, we define motifs as the functional groups that are critical in determining a molecule's properties. For motif extraction from molecular graphs, we adopt a molecule fragmentation method that leverages a retrosynthesis-based algorithm. This approach allows for the identification of semantically meaningful motifs. A motif vocabulary is constructed after processing the entire molecule dataset. This process yields a dedicated feature set for the functional groups, $\mathbf{H}^{(\text{Motif})}$. Both the hierarchical structural features $\{\mathbf{H}^{(l)}\}$ and the motif features $\mathbf{H}^{(\text{Motif})}$ are then passed collectively to our HAP module.

\subsection{Hierarchical Adaptive Projector}
HAP is designed to process the hierarchical feature from the GNN encoder. Its goal is to align the hierarchical feature with downstream tasks. To achieve this, HAP employs a Mixture-of-Experts (MoE) architecture with an adaptively selection mechanism and two specialist projectors.

\paragraph{Specialist Projectors.} The two projectors are designed to handle different types of information:
\begin{enumerate}
    \item \textbf{Cross Attention Projector ($E_{\text{Attn}}$):} This projector is a standard multi-head cross-attention layer implemented in the form of a learnable cross-attention mechanism, where a fixed number of query tokens are learned to attend over node representations. It computes all-pairs interactions among nodes, making it highly effective at identifying globally salient patterns and forming a holistic view of the graph. 
    \item \textbf{Structure-Aware Graph-Mamba Projector ($E_{\text{Mamba}}$):} This projector leverages state-space models to capture sequential and contextual information. Given node features $\mathbf{H}^{(l)}$, we first serialize them using a graph node sorting strategy that serializes atoms by considering both molecular fragments and node degrees, creating an input sequence optimized for Mamba's context-aware reasoning. Then it processes this sequence using a GraphSSM (GSSM) mechanism that integrates structural information, such as adjacency and distance matrices.
\end{enumerate}

\paragraph{Dynamic Gating and Projector Selection.}
For each layer's feature representation $\mathbf{H}^{(l)}$, a linear layer gating network $G_{h}$ determines which projector should process the input. The gating network takes each representation as input and outputs a probability for selecting the projectors.
\begin{equation}
    p^{(l)} = G_{h}(\mathbf{H}^{(l)}) .
\end{equation}
The projector with the highest probability is chosen to process the entire input $\mathbf{H}^{(l)}$. Let $k = \arg\max(p^{(l)})$, then the output of the HAP for layer $l$ is:
\begin{equation}
    \mathbf{Z}^{(l)} = E_k(\mathbf{H}^{(l)}),
\end{equation}
where $E_k$ is the selected projector ($E_{\text{Attn}}$ or $E_{\text{Mamba}}$). This sparse activation mechanism forces the model to make a definitive choice about the best processing strategy for each hierarchy level. This dynamic, layer-by-layer selection allows HAP to adaptively leverage cross-attention and Mamba for hierarchical features.

\subsection{Source-Aware Fusion Module}
Molecular identity is defined by more than just atom-level hierarchical features. Other information sources, such as functional group motifs or global physicochemical properties, are also crucial. To adaptively assemble all this information, we design SAF module.

The input to this module is the concatenated features $\mathbf{Z} = \{\mathbf{Z}^{(1)}, \mathbf{Z}^{(2)}, \dots, \mathbf{Z}^{(L)},\mathbf{Z}^{(\text{Motif})} \} $ from the HAP. For each token $\mathbf{z}_j$ in $\mathbf{Z}$, a linear layer gating network $G_{s}$  will routes each token to a combination of experts. Instead of activating all experts, we employ a sparse Top-2 routing strategy. For a sequence of $M$ feature tokens, we select the two experts with the highest probability for each token $\mathbf{z}_j$, . 

\begin{equation}
I_j = \text{TopK}(G_s(\mathbf{z}_j), k=2).
\end{equation}

The SAF module contains $N$ expert networks, $\{E_1, E_2, \dots, E_N\}$, where each expert $E_i$ is the i-$th$ independent MLP. The final representation for the token $\mathbf{z}_j$, denoted as $\mathbf{y}_j$, is computed by summing the outputs of its two selected experts:
\begin{equation}
\mathbf{y}_j = \sum_{k \in I_j} E_i(\mathbf{z}_j).
\end{equation}

This process is applied independently to every token in the input set $\mathbf{Z}$, obtaining a sequence of processed tokens $\mathbf{Y} = \{\mathbf{y}_1, \mathbf{y}_2, \dots, \mathbf{y}_M\}$. This expressive representation $\mathbf{Y}$, which now compactly encodes the adaptively fused hierarchical and semantic information through the sparse MoE mechanism, serves as the final input to the LLM for various downstream molecular tasks.

\subsection{Instruction-following Respons of LLM}
In the final stage, LLM is employed to generate the instruction-following response. The LLM is conditioned on three distinct input: the molecule's Smiles string $\mathbf{T}_{\text{Smiles}}$, the comprehensive graph representation $\mathbf{Y}$ from SAF module, and a task instruction $\mathbf{T}_{\text{Instruct}}$.

These three input form the complete inputs context. The $\text{LLM}$, processes this context to produce the final textual output $\mathbf{T}_{\text{out}}$:
\begin{equation}
\mathbf{T}_{\text{out}} = \text{LLM}(\mathbf{T}_{\text{SMILES}}, \mathbf{Y}, \mathbf{T}_{\text{Instruct}}).
\end{equation}
This allows the LLM to guide its final response generation in both the smiles and graph structural representations of the molecule, guided by the user's specific instructions.

\section{Experiments}
\subsection{Experimental Settings}
To evaluate the efficacy of our proposed method, we evaluate the model on three public tasks: 1) molecule description generation, 2) IUPAC name prediction, 3) molecule property prediction. We conduct experiments under two major settings: generalist and specialist models. In the generalist setting, a single model is trained to handle all three tasks, whereas in the specialist setting, we train a dedicated model for each downstream task. Our experiments cover six datasets, including the Mol-Instructions, which consists of five datasets (like IUPAC and PubChem), and ChEBI-20.

\subsection{Evaluation Metrics}
To comprehensively assess the performance of our proposed model across different tasks, we employ a set of standard evaluation metrics for each task category. We evaluate the quality of generated text for molecular captioning and IUPAC name prediction, and the accuracy of numerical predictions for the property question-answering task.

\subsubsection{Text Generation Quality}
For tasks requiring the generation of natural language or structured names, we use two metrics: 1) \textbf{BLEU} \cite{papineni2002bleu}: The Bilingual Evaluation Understudy score measures the n-gram precision between the generated text and a reference text. In our experiments, we report BLEU-4, which considers n-grams up to length four. 2) \textbf{METEOR} \cite{banerjee2005meteor}: The Metric for Evaluation of Translation with Explicit ORdering provides a more nuanced assessment than BLEU by considering synonymy, stemming, and word order.

\subsubsection{Property Prediction Accuracy}
For the molecular properties prediction task, we evaluate the model's accuracy using Mean Absolute Error (MAE). This metric calculates the average magnitude of the errors between the predicted values and the ground-truth values, providing a direct measure of prediction accuracy. 


\begin{table*}[ht]
\centering

\begin{tabular}{@{}llcccccc@{}}
\toprule
\multirow{2}{*}{\textbf{Model}} & \multirow{2}{*}{\textbf{LLM}} & \textbf{Mol. Inst.} & \multicolumn{2}{c}{\textbf{Molecule Description}} & \multicolumn{2}{c}{\textbf{IUPAC Prediction}} & \textbf{Property pred.} \\
\cmidrule(lr){4-5} \cmidrule(lr){6-7}
& & \textbf{tuned} & \textbf{BLEU (\(\uparrow\))} & \textbf{METEOR (\(\uparrow\))} & \textbf{BLEU (\(\uparrow\))} & \textbf{METEOR (\(\uparrow\))} & \textbf{MAE (\(\downarrow\))} \\
\midrule
GPT-3.5 & GPT-3.5 & \ding{55} & 10.4 & 27.9 & 34.7 & 44.9 & 0.0529 \\
GPT-3.5 (ICL) & GPT-3.5 & \ding{55} & 14.1 & 40.5 & 40.8 & 62.0 & 0.0364 \\
GPT-4 & GPT-4 & \ding{55} & 10.7 & 21.2 & 36.5 & 48.1 & 0.1007 \\
GPT-4 (ICL) & GPT-4 & \ding{55} & 12.7 & 39.9 & 39.9 & 47.4 & 0.0185 \\
Galactica\textdagger & Galactica & \ding{55} & 1.7 & 22.4 & -- & -- & 0.5680 \\
Text+Chem T5\textdagger & T5-Base & \ding{55} & 3.6 & 13.9 & -- & -- & -- \\
\midrule
LLaMA2 & LLaMA2-7B & \ding{55} & 0.0 & 7.3 & 23.3 & 29.8 & N/A* \\
Mol-Instructions\textdagger & LLaMA2-7B & \ding{51} & 14.3 & 25.4 & -- & -- & 0.0121 \\
LLaMo & LLaMA2-7B & \ding{51} & \underline{37.9} & \underline{67.1} & \underline{61.1} & \underline{74.1} & \underline{0.0061} \\

\textbf{HSA-Net (Ours)} & \textbf{LLaMA2-7B} & \textbf{\ding{51}} & \textbf{43.5} & \textbf{72.1} & \textbf{65.4} & \textbf{78.9} & \textbf{0.0049} \\
\bottomrule
\end{tabular}
\caption{Evaluation results (\%) of generalist models across three tasks: molecule description generation, IUPAC name prediction and molecular property regression. Models labeled ``Mol. Inst. tuned" indicate those trained with molecular instruction tuning. * MAE for LLaMA2 is not reported as it fails to generate valid numerical outputs.
† Results indicated with † are reproduced from Mol-Instruction \cite{fang2023mol}. -- indicates that the result could not be reproduced on the corresponding dataset.}
\label{tab:generalist_performance}
\end{table*}

\begin{table*}[ht]
\centering

\begin{tabular}{@{}llc cc cc c@{}}
\toprule
\multirow{2}{*}{\textbf{Model}} & \multirow{2}{*}{\textbf{LLM}} & \multirow{2}{*}{\textbf{Training type}} & \multicolumn{2}{c}{\textbf{PubChem324kV2}} & \multicolumn{2}{c}{\textbf{ChEBI-20}} & {\textbf{IUPAC}} \\
\cmidrule(lr){4-5} \cmidrule(lr){6-7}
& & & \textbf{BLEU} & \textbf{METEOR} & \textbf{BLEU} & \textbf{METEOR} & \textbf{METEOR} \\
\midrule
MolT5-Small & T5-Small & Full FT & 9.2 & 19.3 & 37.1 & 54.2 & 43.2 \\
MolT5-Base & T5-Base & Full FT & 22.5 & 34.3 & 38.7 & 53.9 & 49.7 \\
MolT5-Large & T5-Large & Full FT & 27.2 & 41.2 & 39.1 & 55.0 & 50.1 \\
\midrule
MoMu-Small & T5-Small & Full FT & 13.1 & 20.6 & 44.5 & 55.7 & -- \\
MoMu-Base & T5-Base & Full FT & 23.7 & 35.4 & 46.2 & 57.6 & -- \\
MoMu-Large & T5-Large & Full FT & 25.4 & 38.7 & \underline{51.5} & 59.7 & -- \\
MolCA, Galac$_{125\text{M}}$ & Galactica-125M & Full FT & 26.8 & 39.1 & 42.6 & 54.4 & 62.1 \\
MolCA, Galac$_{1.3\text{B}}$ & Galactica-1.3B & LoRA & 30.3 & 45.6 & 44.2 & 57.1 & 65.4 \\
BioT5+ & T5-Base & Full FT & -- & -- & 39.2 & 51.1 & -- \\
XMolCap & T5-Base & Full FT & -- & -- & 51.1 & \underline{64.9} & -- \\
LLaMo, Galac$_{1.3\text{B}}$ & Galactica-1.3B & LoRA & \underline{35.2} & \underline{52.1} & 49.9 & 64.8 & \underline{70.1} \\
\midrule
\textbf{HSA-Net (Ours)} & Galactica-1.3B & LoRA & \textbf{37.5} & \textbf{63.5} & \textbf{52.2} & \textbf{69.9} & \textbf{72.8} \\
\bottomrule
\end{tabular}
\caption{Performance (\%) of specialist models on molecule captioning with the PubChem324k and
ChEBI-20 datasets and IUPAC name prediction. Full FT denotes full parameter fine-tuning. -- indicates that the result is irreproducible on the dataset.}
\label{tab:model_performance_comparison}
\end{table*}

\subsubsection{Implementation Details}
For the generalist models, our HSA-Net is built upon the \textbf{LLaMA-2-7b-chat} \cite{touvron2023llama} LLM to ensure a fair comparison with instruction-tuned baselines like Mol-Instructions dataset \cite{fang2023mol}, which contains five public datasets. For the specialist models, we build our HSA-Net on \textbf{Galactica 1.3B} \cite{taylor2022galactica} to maintain a fair comparison with specialist baselines such as MolCA \cite{liu2023molca}.

To train the generalist variant of our HSA-Net, we adopt the two-stage pipeline: 1) \textbf{Representation Alignment.} We use the training split of the molecular description generation dataset from Mol-Instruction \cite{fang2023mol} to align our graph encoder and the HSA module with the frozen LLM. 2) \textbf{Instruction-tuning.} Instruction tuning is conducted to improve alignment with task-specific objectives. The graph encoder is kept frozen, while the HSA modules remain trainable. The LLM is updated via LoRA-based parameter-efficient fine-tuning to reduce computational overhead. The tuning process leverages a mixture of instruction datasets, including molecular description and property prediction data from Mol-Instructions \cite{fang2023mol}, the IUPAC naming dataset from \cite{liu2023molca}, and GPT-4-generated multi-turn instruction-following data. This diverse dataset composition is intended to facilitate better generalization across molecular tasks and representations.

To train the specialist variant of HSA-Net, we follow MolCA \cite{liu2023molca} setting. The model is pre-trained on a split of PubChem324k \cite{liu2023molca} in Stage 1, and then fine-tuned for specific downstream task in Stage 2. Further implementation details are available in Appendix A.1.

\begin{figure*}[h!]
    \centering
    \begin{subfigure}[b]{0.32\textwidth}
        \centering
        \includegraphics[width=\textwidth]{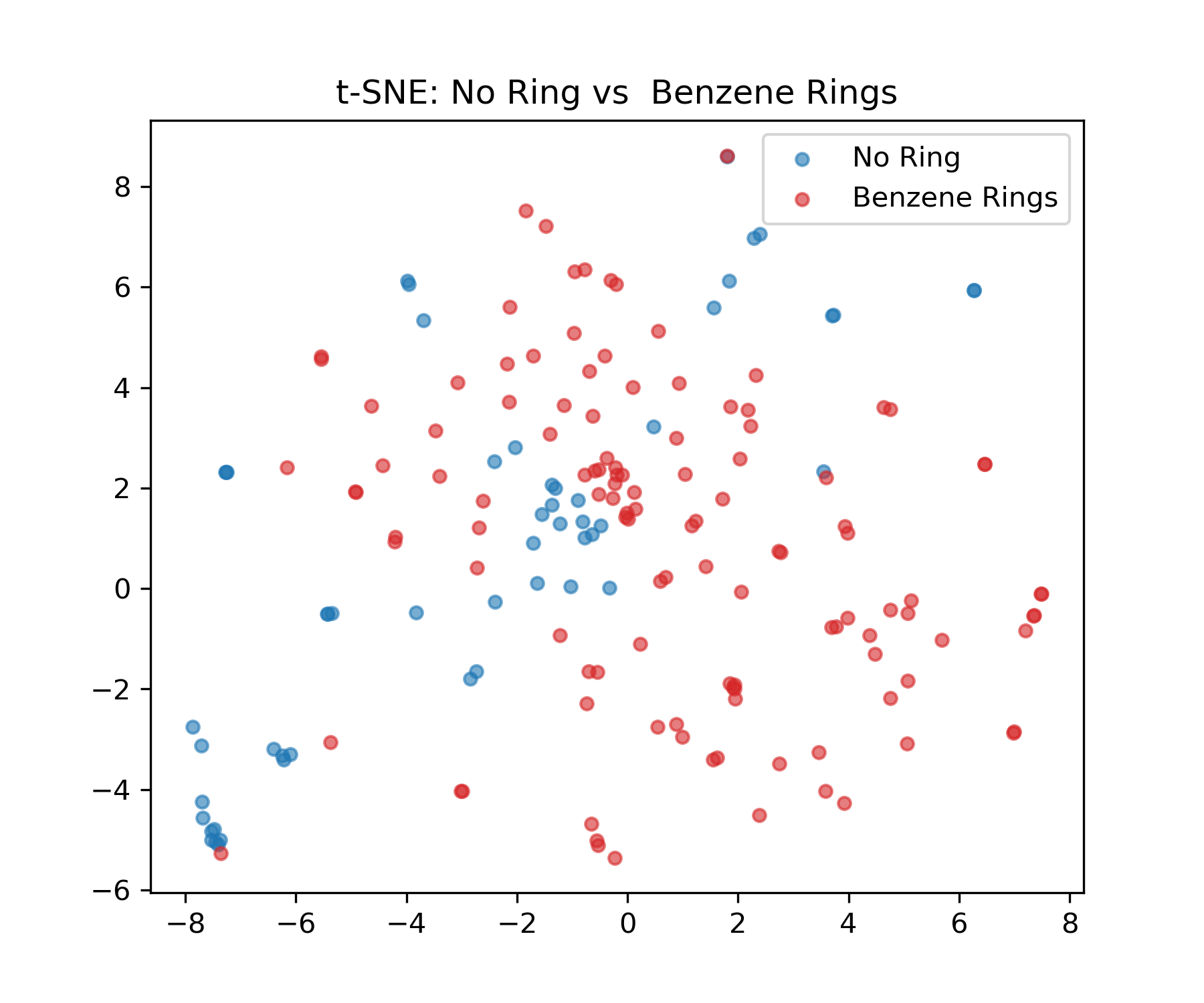}
        \caption{ Attention Only }
        \label{fig:attention_only}
    \end{subfigure} 
    \hfill
    \begin{subfigure}[b]{0.32\textwidth}
        \centering
        \includegraphics[width=\textwidth]{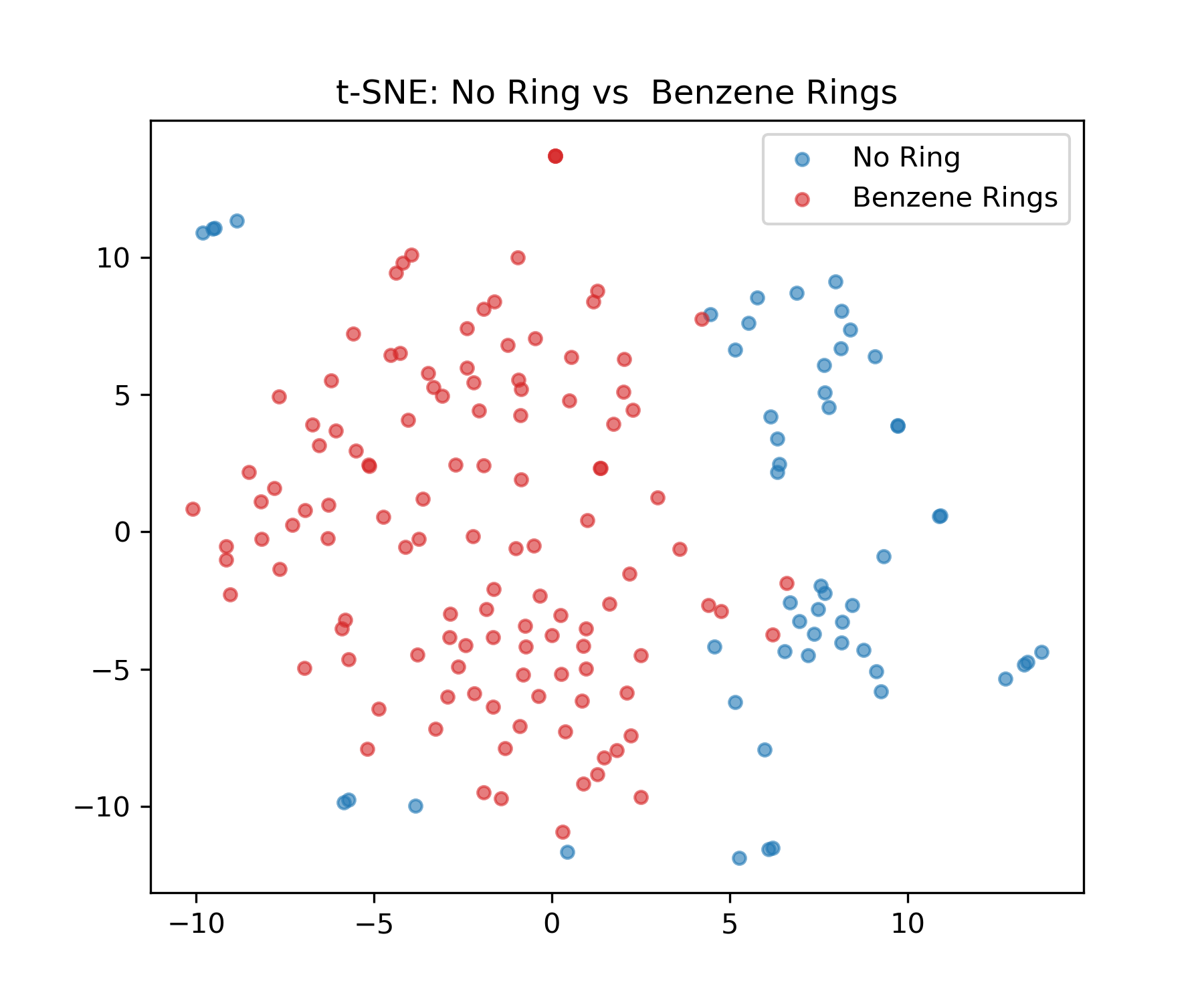}
        \caption{ Graph-Mamba Only }
        \label{fig:mamba_only}
    \end{subfigure}
    \hfill
        \begin{subfigure}[b]{0.32\textwidth}
        \centering
        \includegraphics[width=\textwidth]{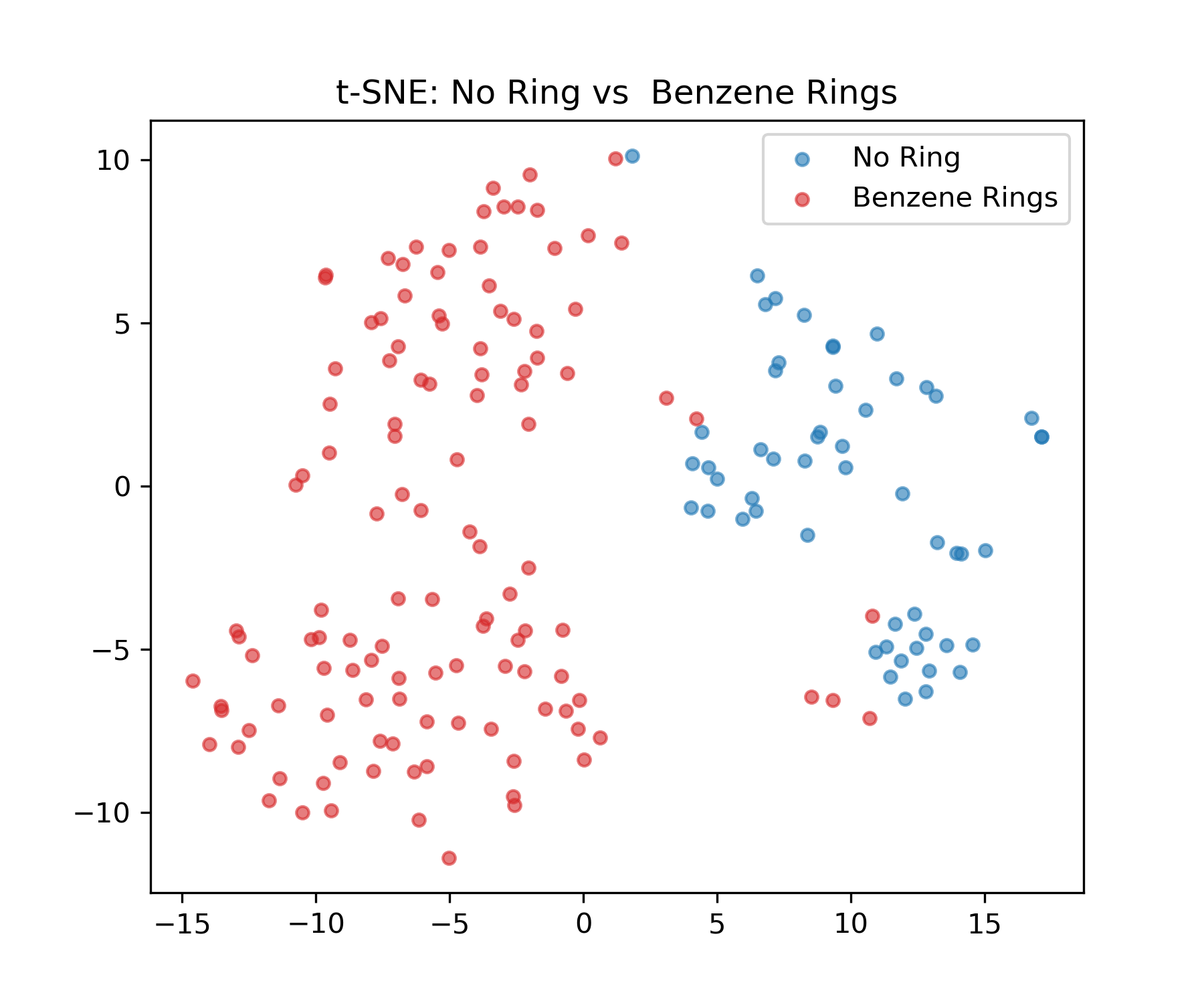}
        \caption{HSA-Net (Ours)}
        \label{fig:hsa_net_full}
    \end{subfigure}
    \caption{
        t-SNE visualization of final layer molecule-level features for molecules with (red) and without (blue) a benzene ring. 
        (a) The ``Attention Only" baseline suffers from severe feature overlap, demonstrating over-smoothing and a loss of discriminative ability.
        (b) The ``Graph-Mamba Only" projector maintains some separation, but still has some overlap. 
        (c) The full HSA-Net model shows distinct, well-separated clusters, indicating a highly discriminative representation. 
    }
    \label{fig:tsne}
\end{figure*}

\subsubsection{Baselines}
For the generalist models, we compare our HSA-Net with (1) general-purpose LLM-based models, including Galactica \cite{taylor2022galactica}, LLaMA2-7B \cite{touvron2023llama}, GPT-3.5, and GPT-4; (2) molecule-specialized LLM, Text+Chem T5 \cite{christofidellis2023unifying}; and (3) molecule-instruction-tuned generalist models, Mol-Instructions \cite{fang2023mol} and LLaMo(with LLaMA2) \cite{park2024llamo}. Since GPT-3.5 and GPT-4 struggle with these tasks in a zero-shot setting, we also report their performance with 4-shot in-context learning (ICL). For the specialist models, we use single-task specialist molecule-language models as baselines, including MolT5 \cite{edwards2022translation}, MoMu \cite{su2022molecular}, MolCA \cite{liu2023molca}, BioT5+ \cite{pei2024biot5+}, LLaMo(with Galactica) \cite{park2024llamo} and XMolCap \cite{tran2025xmolcap}.

\begin{table}[t!]
\centering

\begin{tabular}{ccc|cc}
\toprule
Attention & Mamba & SAF & BLEU ↑ & METEOR ↑ \\
\midrule
\ding{51} & \ding{55} & \ding{55} & 37.9 & 67.1 \\  
\ding{55} & \ding{51} & \ding{55} & 37.5 & 67.9 \\  
\ding{51} & \ding{51} & \ding{55} & 38.8 & 68.5 \\  
\ding{51} & \ding{55} & \ding{51} & 40.0 & 69.2 \\  
\ding{55} & \ding{51} & \ding{51} & 39.4 & 69.9 \\  
\ding{51} & \ding{51} & \ding{51} & \textbf{43.5} & \textbf{72.1} \\  
\bottomrule
\end{tabular}
\caption{Ablation study of HSA-Net variants with different architectural components.}
\label{tab:ablation_bleu_meteor}
\end{table}

\begin{figure}[h]
    \centering
    \includegraphics[width=0.95\columnwidth]{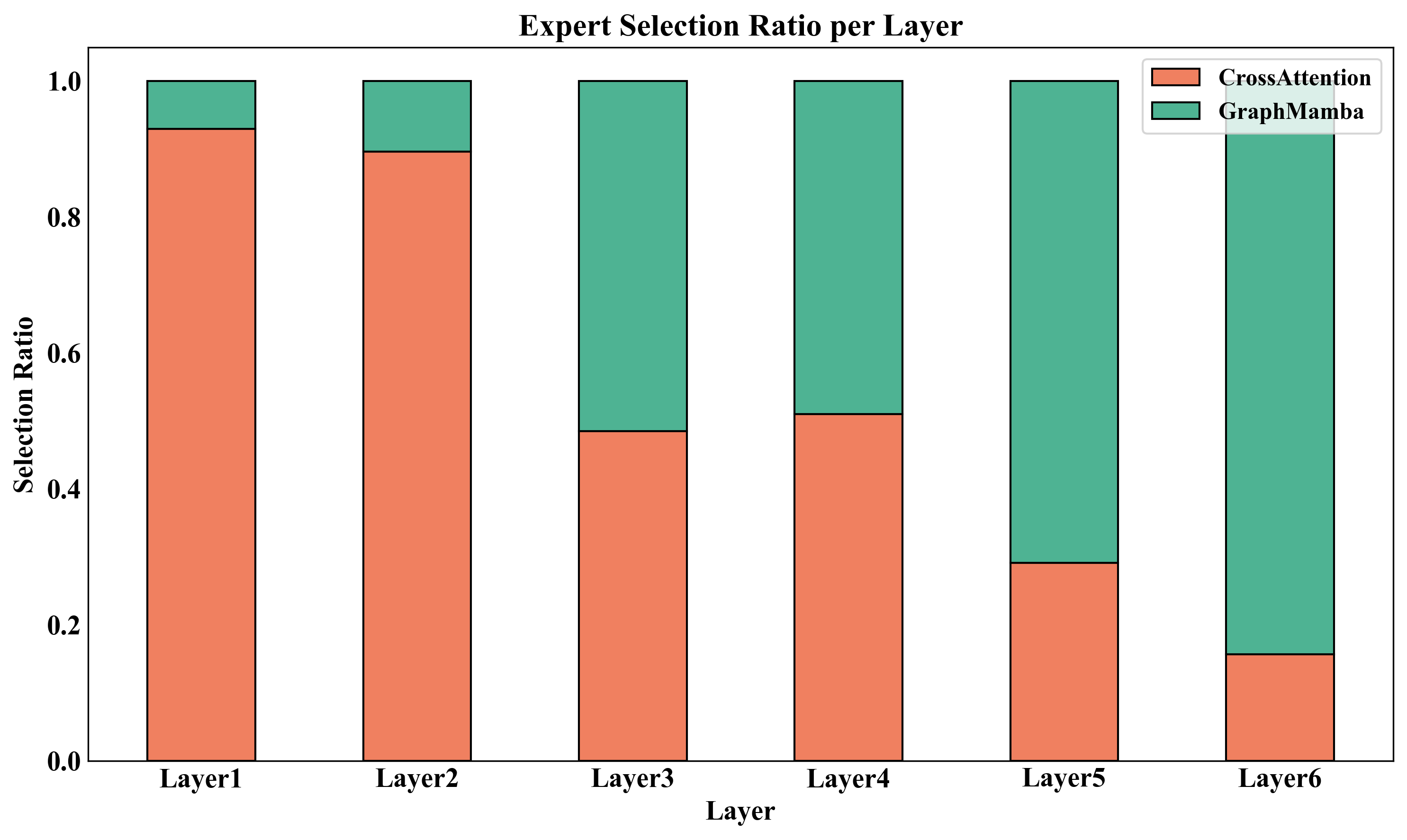} 
    \caption{Visualization of molecule feature distributions using t-SNE. The plots compare the final graph-level representations of a molecule dataset at GNN Layer 1 and Layer 6, generated by Cross-Attention and Graph-Mamba.}
    \label{fig:gating}
\end{figure}

\subsection{Quantitative Results and Analysis}
\label{sec:quantitative}
\subsubsection{Comparison with SOTA Methods}
Table \ref{tab:generalist_performance} and Table \ref{tab:model_performance_comparison} provides a detailed comparison of HSA-Net against baseline and SOTA methods. HSA-Net demonstrates remarkable effectiveness, achieving the highest performance across all evaluated tasks in both generalist and specialist settings. As shown in Table \ref{tab:generalist_performance}, for generalist models, our HSA-Net achieves a BLEU score of 43.5 and a METEOR score of 72.1 in molecule description, significantly outperforming the next best model, LLaMo. This represents an improvement of 5.6 \% in BLEU and 5.0 \% in METEOR over SOTA. The performance in IUPAC name prediction is also superior, where HSA-Net's scores of 65.4 in BLEU and 78.9 in METEOR surpass the runner-up by 4.3 and 4.8 \%, respectively. In property prediction, HSA-Net achieves a MAE value of 0.0049, improving the performance by nearly 20 \% compared to LLaMo's 0.0061. 

Similarly, in the specialist model comparisons shown in Table 2, HSA-Net consistently surpasses its competitors. For instance, in IUPAC name prediction, it achieves a METEOR score of 72.8, improving upon the next-best score of 70.1 from LLaMo by 2.7 \%. The consistent and significant margins of improvement across diverse tasks and datasets underscore a clear trend: the architectural innovations within HSA-Net, which effectively resolve the global-local trade-off in molecular feature projection, enable it to attain a new SOTA in molecular representation learning.

\begin{figure*}[h!]
\centering
\includegraphics[width=0.4\textwidth]{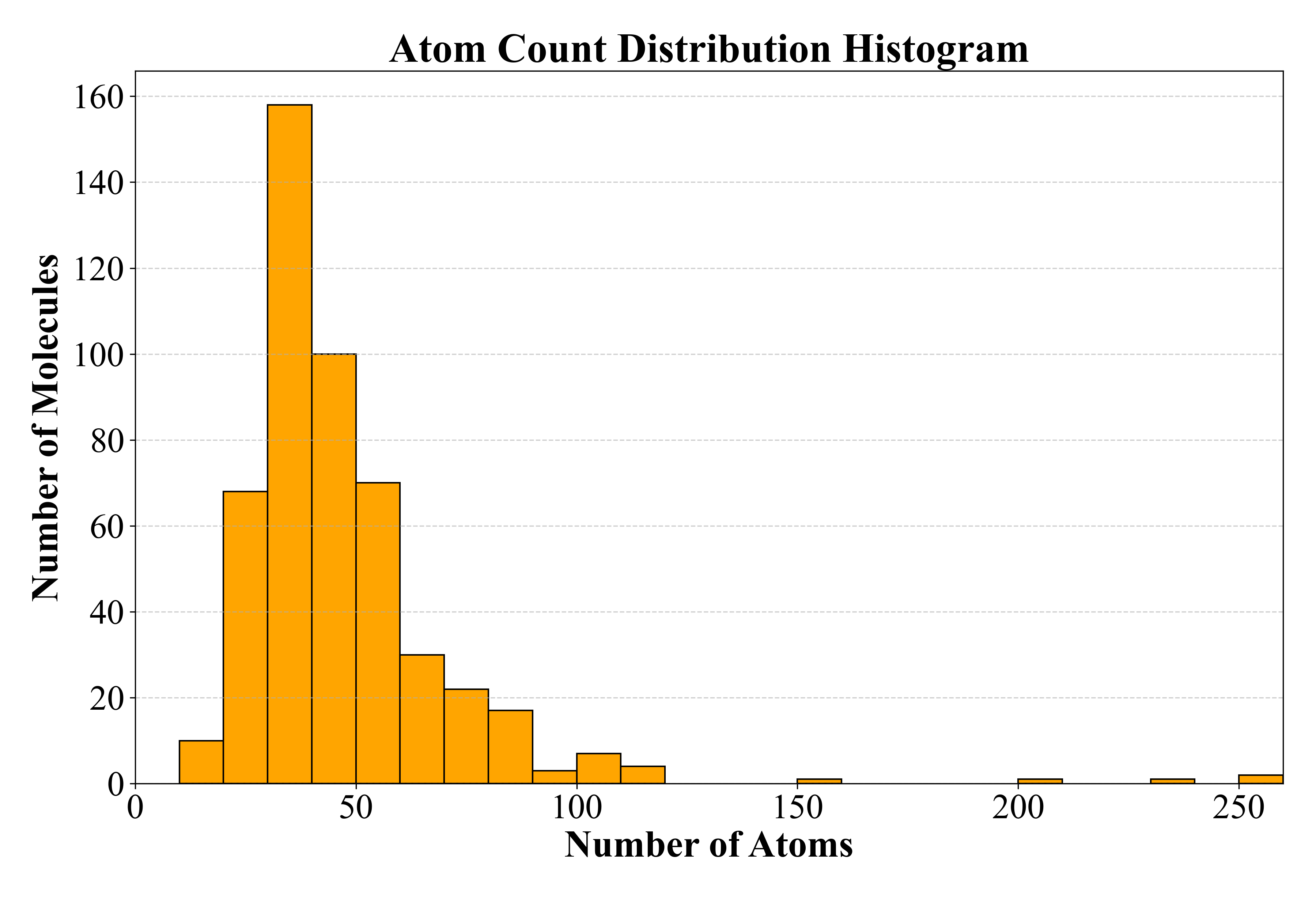}
\includegraphics[width=0.54\textwidth]{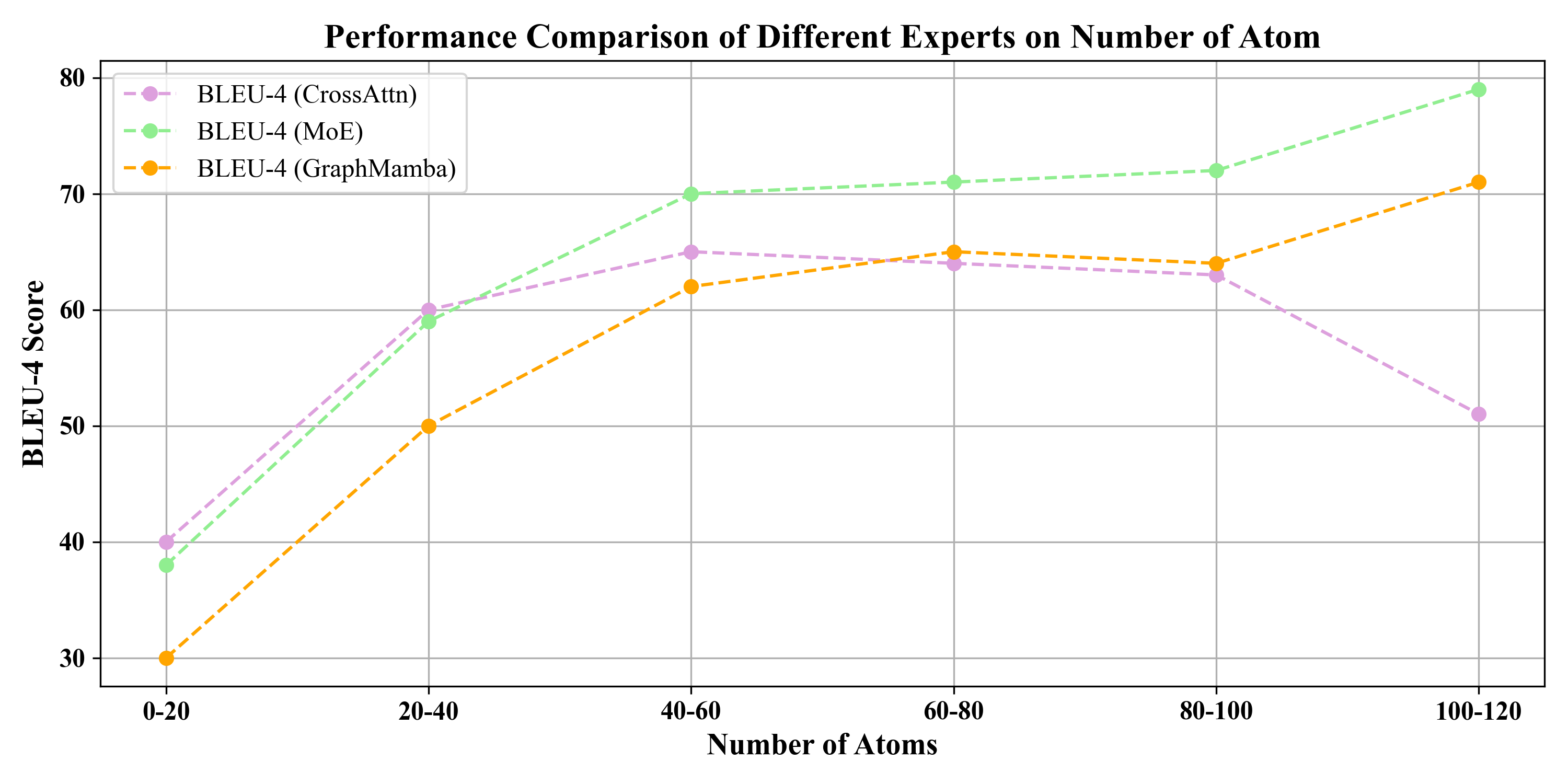}
\caption{\textbf{Analysis of Model Performance by Molecular Size.} \textbf{The left figure:} The atom count distribution of our test dataset, showing that the majority of molecules are relatively small (fewer than 60 atoms). \textbf{The right figure:} A comparison of evaluation metrics between our HSA-Net and a standard Cross-Attention baseline, segmented by molecular size. The HSA-Net (dashed lines) demonstrates significantly better performance on larger molecules (60+ atoms), indicating its capability to model long-range dependencies and complex molecule structure.}
\label{fig:size_analysis}
\end{figure*}


\subsubsection{Ablation Study}
To assess the contribution of each module in HSA-Net, we conduct an ablation study, as shown in Table~\ref{tab:ablation_bleu_meteor}. For ablation experiments, we train on the basis of the generalist model with the same settings. The table enumerates combinations of three key modules:
1) cross-attention projector. 2) Graph-Mamba projector. 3) SAF module. When SAF is disabled, it is replaced with an MLP that directly performs token fusion.

The results reveal several trends: First, activating both cross-attention and Graph-Mamba leads to better performance than using either path alone, suggesting their complementarity in capturing both local and global structural information. Second, the inclusion of SAF consistently improves performance across different backbone configurations, validating the value of adaptive expert fusion. Finally, the full model, which includes all three components, achieves the highest BLEU (43.5) and METEOR (72.1) scores, demonstrating the effectiveness of combining diverse pathways with adaptive routing in HSA-Net.


\subsection{Qualitative Results and Analysis}
\label{sec:qualitative}

\subsubsection{Visualization of Feature Representation}
To qualitatively analyse how HSA-Net mitigates over-smoothing and obtains diverse multi-level representations, we use t-SNE \cite{maaten2008visualizing} to visualize the molecule-level feature space of the final feature. As shown in Figure \ref{fig:tsne}, we compare the feature distributions for molecules containing and not containing a benzene ring to investigate the model ability for modeling complex molecule structure. For the ``Attention Only" projector, the feature clusters for the two classes show significant overlap, indicating severe over-smoothing and loss of discriminative ability. In contrast, our full HSA-Net model produces two distinct, well-separated clusters. This qualitatively confirms that our hierarchical approach, by switching to the structure-aware Graph-Mamba projector in deep layers, effectively preserves feature diversity and generates a much more expressive final representation that aligns with down-stream tasks.

\subsubsection{Analysis of HAP Gating Decisions}
To further investigate the mechanism of projector selection of our model, we analyze the gating decisions made by the HAP module across different GNN layers. As depicted in Figure \ref{fig:gating}, we plot the average ratio assigned to the Graph-Mamba projector for each GNN layer. The results reveal a clear trend: for shallow layers (1-2), the model assigns low ratio to the Mamba projector, preferring the cross-attention projector. As the layer depth increases, the ratio shifts towards the Mamba projector. This provides direct evidence that our framework has learned the hierarchical projection model selection. It learns to use cross-attention for low-level, local features and Mamba for high-level, abstract structural features, providing effective solution for adaptive feature projection.

\subsubsection{Analysis of Molecule Length} To investigate the effectiveness of our architecture on different molecule size, we analyzed the performance trends of our HSA-Net against a standard cross-attention baseline, segmented by atom count. We focused our analysis on molecules within the 0-120 atom range, which informed by our dataset's atom count distribution, which shows a high concentration of molecules in this scope and a sparse population beyond it. As illustrated in Figure \ref{fig:size_analysis}, a distinct trend reveals that the HSA-Net demonstrates superior performance for larger molecules. While both models perform competitively on smaller molecules (less than 40 atoms), the BLEU and METEOR scores of HSA-Net consistently surpass the cross-attention for longer molecules, with the performance gap larger as the atom count increases. This result validate that our HSA-Net architecture's capability to effectively model the complex, long-range dependencies inherent in larger molecular structures.

\section{Conclusion}
In this work, we introduce HSA-Net, a novel framework designed to resolve the global-local trade-off of feature projection in molecular representation learning. HSA-Net contains two modules: the HAP module dynamically switches between a cross-attention projector and a structure-aware Graph-Mamba projector to process features from different GNN layers, effectively mitigating the over-smoothing problem. Subsequently, the SAF module adaptively integrates these hierarchical features, obtaining a comprehensive and expressive final representation. Extensive experiments demonstrate that our proposed HSA-Net quantitatively and qualitatively outperforms current SOTA methods on down-stream tasks. Further ablation studies and visualizations validate the effectiveness of our approach. For future work, investigating a broader range of molecular types, such as proteins, and more complex biological tasks is a worthwhile direction.

\section{Acknowledgement}
This work was supported by the National Natural Science Foundation of China (No. 62471420), GuangDong Basic and Applied Basic Research Foundation (2025A1515012296), and CCF-Tencent Rhino-Bird Open Research Fund.


\bigskip
\bibliography{aaai2026}

\begin{thebibliography}{24}
\providecommand{\natexlab}[1]{#1}

\bibitem[{Achiam et~al.(2023)Achiam, Adler, Agarwal, Ahmad, Akkaya, Aleman, Almeida, Altenschmidt, Altman, Anadkat et~al.}]{achiam2023gpt}
Achiam, J.; Adler, S.; Agarwal, S.; Ahmad, L.; Akkaya, I.; Aleman, F.~L.; Almeida, D.; Altenschmidt, J.; Altman, S.; Anadkat, S.; et~al. 2023.
\newblock Gpt-4 technical report.
\newblock \emph{arXiv preprint arXiv:2303.08774}.

\bibitem[{Banerjee and Lavie(2005)}]{banerjee2005meteor}
Banerjee, S.; and Lavie, A. 2005.
\newblock METEOR: An automatic metric for MT evaluation with improved correlation with human judgments.
\newblock In \emph{Proceedings of the acl workshop on intrinsic and extrinsic evaluation measures for machine translation and/or summarization}, 65--72.

\bibitem[{Christofidellis et~al.(2023)Christofidellis, Giannone, Born, Winther, Laino, and Manica}]{christofidellis2023unifying}
Christofidellis, D.; Giannone, G.; Born, J.; Winther, O.; Laino, T.; and Manica, M. 2023.
\newblock Unifying molecular and textual representations via multi-task language modelling.
\newblock In \emph{International Conference on Machine Learning}, 6140--6157. PMLR.

\bibitem[{Edwards et~al.(2022)Edwards, Lai, Ros, Honke, Cho, and Ji}]{edwards2022translation}
Edwards, C.; Lai, T.; Ros, K.; Honke, G.; Cho, K.; and Ji, H. 2022.
\newblock Translation between molecules and natural language.
\newblock \emph{arXiv preprint arXiv:2204.11817}.

\bibitem[{Fang et~al.(2023)Fang, Liang, Zhang, Liu, Huang, Chen, Fan, and Chen}]{fang2023mol}
Fang, Y.; Liang, X.; Zhang, N.; Liu, K.; Huang, R.; Chen, Z.; Fan, X.; and Chen, H. 2023.
\newblock Mol-instructions: A large-scale biomolecular instruction dataset for large language models.
\newblock \emph{arXiv preprint arXiv:2306.08018}.

\bibitem[{Gu and Dao(2023)}]{gu2023mamba}
Gu, A.; and Dao, T. 2023.
\newblock Mamba: Linear-time sequence modeling with selective state spaces.
\newblock \emph{arXiv preprint arXiv:2312.00752}.

\bibitem[{Hu et~al.(2025)Hu, Guo, Si, Liu, Diao, Zhang, Zhou, and Wang}]{hu2025mol}
Hu, J.; Guo, D.; Si, Z.; Liu, D.; Diao, Y.; Zhang, J.; Zhou, J.; and Wang, M. 2025.
\newblock MOL-Mamba: Enhancing Molecular Representation with Structural \& Electronic Insights.
\newblock In \emph{Proceedings of the AAAI Conference on Artificial Intelligence}, volume~39, 317--325.

\bibitem[{Ji et~al.(2022)Ji, Shi, Lu, Li, and Yang}]{ji2022relmole}
Ji, Z.; Shi, R.; Lu, J.; Li, F.; and Yang, Y. 2022.
\newblock ReLMole: Molecular Representation Learning Based on Two-Level Graph Similarities.
\newblock \emph{Journal of chemical information and modeling}, 62.

\bibitem[{Kipf and Welling(2016)}]{kipf2016semi}
Kipf, T.~N.; and Welling, M. 2016.
\newblock Semi-supervised classification with graph convolutional networks.
\newblock \emph{arXiv preprint arXiv:1609.02907}.

\bibitem[{Li, Han, and Wu(2018)}]{li2018deeper}
Li, Q.; Han, Z.; and Wu, X.-M. 2018.
\newblock Deeper insights into graph convolutional networks for semi-supervised learning.
\newblock In \emph{Proceedings of the AAAI conference on artificial intelligence}, volume~32.

\bibitem[{Liu et~al.(2023)Liu, Li, Luo, Fei, Cao, Kawaguchi, Wang, and Chua}]{liu2023molca}
Liu, Z.; Li, S.; Luo, Y.; Fei, H.; Cao, Y.; Kawaguchi, K.; Wang, X.; and Chua, T.-S. 2023.
\newblock Molca: Molecular graph-language modeling with cross-modal projector and uni-modal adapter.
\newblock \emph{arXiv preprint arXiv:2310.12798}.

\bibitem[{Maaten and Hinton(2008)}]{maaten2008visualizing}
Maaten, L. v.~d.; and Hinton, G. 2008.
\newblock Visualizing data using t-SNE.
\newblock \emph{Journal of machine learning research}, 9(Nov): 2579--2605.

\bibitem[{Papineni et~al.(2002)Papineni, Roukos, Ward, and Zhu}]{papineni2002bleu}
Papineni, K.; Roukos, S.; Ward, T.; and Zhu, W.-J. 2002.
\newblock Bleu: a method for automatic evaluation of machine translation.
\newblock In \emph{Proceedings of the 40th annual meeting of the Association for Computational Linguistics}, 311--318.

\bibitem[{Park et~al.(2024)Park, Bae, Ko, and Kim}]{park2024llamo}
Park, J.; Bae, M.; Ko, D.; and Kim, H.~J. 2024.
\newblock Llamo: Large language model-based molecular graph assistant.
\newblock \emph{Advances in Neural Information Processing Systems}, 37: 131972--132000.

\bibitem[{Pei et~al.(2024)Pei, Wu, Gao, Liang, Fang, Zhu, Xie, Qin, and Yan}]{pei2024biot5+}
Pei, Q.; Wu, L.; Gao, K.; Liang, X.; Fang, Y.; Zhu, J.; Xie, S.; Qin, T.; and Yan, R. 2024.
\newblock Biot5+: Towards generalized biological understanding with iupac integration and multi-task tuning.
\newblock \emph{arXiv preprint arXiv:2402.17810}.

\bibitem[{Rogers and Hahn(2010)}]{rogers2010extended}
Rogers, D.; and Hahn, M. 2010.
\newblock Extended-connectivity fingerprints.
\newblock \emph{Journal of chemical information and modeling}, 50(5): 742--754.

\bibitem[{Scarselli et~al.(2008)Scarselli, Gori, Tsoi, Hagenbuchner, and Monfardini}]{scarselli2008graph}
Scarselli, F.; Gori, M.; Tsoi, A.~C.; Hagenbuchner, M.; and Monfardini, G. 2008.
\newblock The graph neural network model.
\newblock \emph{IEEE transactions on neural networks}, 20(1): 61--80.

\bibitem[{Su et~al.(2022)Su, Du, Yang, Zhou, Li, Rao, Sun, Lu, and Wen}]{su2022molecular}
Su, B.; Du, D.; Yang, Z.; Zhou, Y.; Li, J.; Rao, A.; Sun, H.; Lu, Z.; and Wen, J.-R. 2022.
\newblock A molecular multimodal foundation model associating molecule graphs with natural language.
\newblock \emph{arXiv preprint arXiv:2209.05481}.

\bibitem[{Taylor et~al.(2022)Taylor, Kardas, Cucurull, Scialom, Hartshorn, Saravia, Poulton, Kerkez, and Stojnic}]{taylor2022galactica}
Taylor, R.; Kardas, M.; Cucurull, G.; Scialom, T.; Hartshorn, A.; Saravia, E.; Poulton, A.; Kerkez, V.; and Stojnic, R. 2022.
\newblock Galactica: A large language model for science.
\newblock \emph{arXiv preprint arXiv:2211.09085}.

\bibitem[{Touvron et~al.(2023)Touvron, Martin, Stone, Albert, Almahairi, Babaei, Bashlykov, Batra, Bhargava, Bhosale et~al.}]{touvron2023llama}
Touvron, H.; Martin, L.; Stone, K.; Albert, P.; Almahairi, A.; Babaei, Y.; Bashlykov, N.; Batra, S.; Bhargava, P.; Bhosale, S.; et~al. 2023.
\newblock Llama 2: Open foundation and fine-tuned chat models.
\newblock \emph{arXiv preprint arXiv:2307.09288}.

\bibitem[{Tran et~al.(2025)Tran, Nguyen, Pham, Rakkiyappan, Karki, and Manavalan}]{tran2025xmolcap}
Tran, D.~T.; Nguyen, N. D.~H.; Pham, N.~T.; Rakkiyappan, R.; Karki, R.; and Manavalan, B. 2025.
\newblock XMolCap: Advancing Molecular Captioning through Multimodal Fusion and Explainable Graph Neural Networks.
\newblock \emph{IEEE Journal of Biomedical and Health Informatics}.

\bibitem[{Veli{\v{c}}kovi{\'c} et~al.(2017)Veli{\v{c}}kovi{\'c}, Cucurull, Casanova, Romero, Lio, and Bengio}]{velivckovic2017graph}
Veli{\v{c}}kovi{\'c}, P.; Cucurull, G.; Casanova, A.; Romero, A.; Lio, P.; and Bengio, Y. 2017.
\newblock Graph attention networks.
\newblock \emph{arXiv preprint arXiv:1710.10903}.

\bibitem[{Yang et~al.(2021)Yang, Li, Song, and Cai}]{yang2021deep}
Yang, S.; Li, Z.; Song, G.; and Cai, L. 2021.
\newblock Deep molecular representation learning via fusing physical and chemical information.
\newblock \emph{Advances in neural information processing systems}, 34: 16346--16357.

\bibitem[{Zhao et~al.(2025)Zhao, Chen, Fang, Zhang, and Li}]{zhao2025dual}
Zhao, A.; Chen, Z.; Fang, Z.; Zhang, X.; and Li, J. 2025.
\newblock Dual-Modality Representation Learning for Molecular Property Prediction.
\newblock \emph{arXiv preprint arXiv:2501.06608}.

\end{thebibliography}


\end{document}